\documentclass{article}
\usepackage{spconf,amsmath,graphicx}
\usepackage{caption}
\usepackage{setspace, lipsum}
\usepackage{hyperref}
\usepackage{multirow}
\usepackage{color, xcolor, colortbl}
\usepackage[sorting = none, backend = bibtex, style=numeric-comp]{biblatex}
\AtBeginBibliography{\small}
\usepackage{url}
\usepackage{subcaption}
\usepackage{float}
\usepackage{nccmath}
\usepackage{amssymb,amsthm,bbm}
\usepackage{adjustbox, booktabs}

\usepackage{xcolor}
\hypersetup{
    colorlinks,
    linkcolor={red!50!black},
    citecolor={blue!50!black},
    urlcolor={blue!80!black}
}

\definecolor{darkred}{rgb}{0.7, 0, 0}
\definecolor{darkblue}{rgb}{0.0, 0.0, 0.7}
\definecolor{darkgreen}{rgb}{0, 0.4, 0}

\newcommand{\Du}{\mathcal{D}_u}
\newcommand{\Gen}{\mathcal{G}}
\newcommand{\Dis}{\mathcal{D}}
\newcommand{\Cl}{\mathcal{C}}
\newcommand{\DataPaired}{\mathbf{D}^p}
\newcommand{\DataUnpaired}{\mathbf{D}^{up}}
\title{A Semi-Paired Approach For Label-to-Image Translation}
\name{George Eskandar, Shuai Zhang, Mohamed Abdelsamad, Mark Youssef, Diandian Guo, Bin Yang}

\address{University of Stuttgart, Institute of Signal Processing and System Theory, Stuttgart, Germany}

\begin{document}
\setlength{\abovedisplayskip}{5pt}
\setlength{\belowdisplayskip}{5pt}
\maketitle
\begin{abstract}
Data efficiency, or the ability to generalize from a few labeled data, remains a major challenge in deep learning. Semi-supervised learning has thrived in traditional recognition tasks alleviating the need for large amounts of labeled data, yet it remains understudied in image-to-image translation (I2I) tasks. In this work, we introduce the first semi-supervised (semi-paired) framework for label-to-image translation, a challenging subtask of I2I which generates photorealistic images from semantic label maps. In the semi-paired setting, the model has access to a small set of paired data and a larger set of unpaired images and labels. Instead of using geometrical transformations as a pretext task like previous works, we leverage an input reconstruction task by exploiting the conditional discriminator on the paired data as a reverse generator. We propose a training algorithm for this shared network, and we present a rare classes sampling algorithm to focus on under-represented classes. Experiments on 3 standard benchmarks show that the proposed model outperforms state-of-the-art unsupervised and semi-supervised approaches, as well as some fully supervised approaches while using a much smaller number of paired samples.       
\end{abstract}
\vspace{-1em}
\section{Introduction}
\label{sec:intro}
\vspace{-0.5em}
In recent years, Image-to-Image translation (I2I)~\cite{isola2017image, wang2018high}, the task of mapping an image from a source domain to a target domain,  has gained much momentum by thriving on generative adversarial neural networks (GANs) \cite{Goodfellow2014GenerativeAN}. I2I unleashes a lot of applications, such as style transfer~\cite{Gatys2016}, image inpainting~\cite{isola2017image}, image super-resolution~\cite{ledig2017photo} and semantic image synthesis (SIS) ~\cite{wang2018high, park2019semantic, tan2020rethinking, liu2019learning, Zhu2020SEANIS, schonfeld2021you}. In particular, SIS is a promising data-centric computer graphics approach which seeks to generate photorealistic images from semantic label maps without requiring extensive physics modeling. SIS is a one-to-many mapping: multiple plausible images can be generated from the same semantic map. It spans a wide range of applications from content creation and object manipulation~\cite{park2019semantic} to data augmentation for deep neural networks \cite{palette, urbanstylegan, pragmatic}. 


However, a recurring challenge in SIS and other I2I applications is how to alleviate the need for huge amounts of labeled data for training. Acquiring a large-scale dataset with pairs of images and labels is often resource-intensive and time-consuming. Recent advances in unpaired GANs, which are based on cycle-consistency losses or relationship preservation constraints \cite{zhu2017unpaired, huang2018multimodal, lee2018diverse, park2020cut, fu2019geometry, benaim2017one, taigman2017unsupervised, shrivastava2017learning, bousmalis2017unsupervised, amodio2019travelgan, zhang2019harmonic}, eliminate the need for paired data but lack behind supervised (or paired) models in perceptual quality and semantic alignment. In our previous work~\cite{usiscag, usisicassp}, we introduced USIS, a state-of-the-art unpaired GAN framework for SIS, which achieves competitive fidelity scores (FID) with paired methods but still lack behind in semantic alignment. In real-world scenarios, a small portion of the data can often be labeled without incurring extravagant costs, which raises a valid research question: how can we exploit a small set of labeled data in conjunction with abundant low-cost unpaired data to further improve the quality of generated images? This \textit{semi-paired} strategy (also semi-supervised) has the potential to achieve a good trade-off between cost and performance. Although semi-supervised learning has been extensively studied in visual recognition tasks, it remains largely under-explored in I2I and SIS. There have only been a few works that attempted learning generative tasks in semi-supervised settings. Transformation Consistency Regularization (TCR)~\cite{tcr} is a semi-supervised I2I framework that applies a consistency regularization loss between the model's prediction under different geometrical transformations of the unpaired data. MatchGAN~\cite{matchgan} is a semi-supervised framework to edit faces using attributes (such as "brown hair") as conditional input. SEMIT~\cite{semit} applies a semi-supervised paradigm to a few-shot GAN, and \cite{semiicaiit} employs a weakly-supervised framework for style transfer. Among these frameworks, only TCR can be referred to as semi-paired and can be applied to SIS, as the other frameworks refer to totally different tasks or learning paradigms. Therefore, we argue that there exists a need to develop a semi-supervised paradigm for SIS. While it may seem that adding more unpaired data will definitely boost the model's performance, we show later in the experiments that this is not the case and that semi-supervised generative tasks are non-trivial, requiring a careful model design.  


In this work, we propose the first semi-paired framework for SIS to alleviate the need for collecting labeled data. We argue that leveraging geometrical augmentations like previous works is not enough as SIS is a challenging and under-constrained task where the target image contains more information than the input label. Instead, we propose two novel technical contributions: (1) We leverage the conditional discriminator in the paired setting as a cycle reverse network to train the generator on the unpaired samples. This allows learning a reverse mapping based on the joint probability distribution of images and labels, not just the marginal distribution of images. A new algorithm is proposed to train this shared network. (2) We propose an algorithm to oversample rare classes in the labeled dataset so that more frequent classes do not dominate in the beginning of the training. In this way, we achieve new state-of-the-art results for semi-paired SIS on 3 benchmarks: Cityscapes, COCO-stuff and ADE20K.

\vspace{-1em}
\section{Method}
\label{sec:method}
\vspace{-1em}
In SIS, our goal is to learn a mapping from a semantic label map, $m \in \mathcal{M}$ and a noise vector $z \in \mathcal{N}(0,1)$ to a photorealistic RGB image, $x \in \mathcal{X}$. The labels $m$ are one-hot encoded with $\Cl$ classes. In the semi-paired setting, part of the training data consists of pairs of images and labels, $\DataPaired = \{ (x^p, m^p)_i \}_{i=1}^{N^p}$. The other part consists of samples $\{x_j^{up}\}_{j=1}^{N^{x^{up}}}$ from $\mathcal{X}$ and samples $\{m_j^{up}\}_{j=1}^{N^{m^{up}}}$ from $\mathcal{M}$, where samples from both domains are not aligned. For simplicity, we assume that $N^{x^{up}} = N^{m^{up}} = N^{up} $. We denote the unpaired dataset as $\DataUnpaired = \{x_j^{up}\}_{j=1}^{N^{up}} \bigcup \{m_j^{up}\}_{j=1}^{N^{up}}$. The supervision ratio $r$ is defined as the ratio of paired images to the total number of available images for training, $r = \frac{N^{p}}{N^p + N^{up}}$. Interested readers are referred to~\cite{schonfeld2021you, park2019semantic} for a review of the paired paradigm, and to the USIS~\cite{usisicassp, usiscag} paper for a review of the unpaired paradigm.

\vspace{-1em}
\subsection{The Semi-Paired Framework}
\vspace{-0.5em}

\begin{figure*}[t]
    \centering
    \begin{center}
    \includegraphics[width=0.8\textwidth]{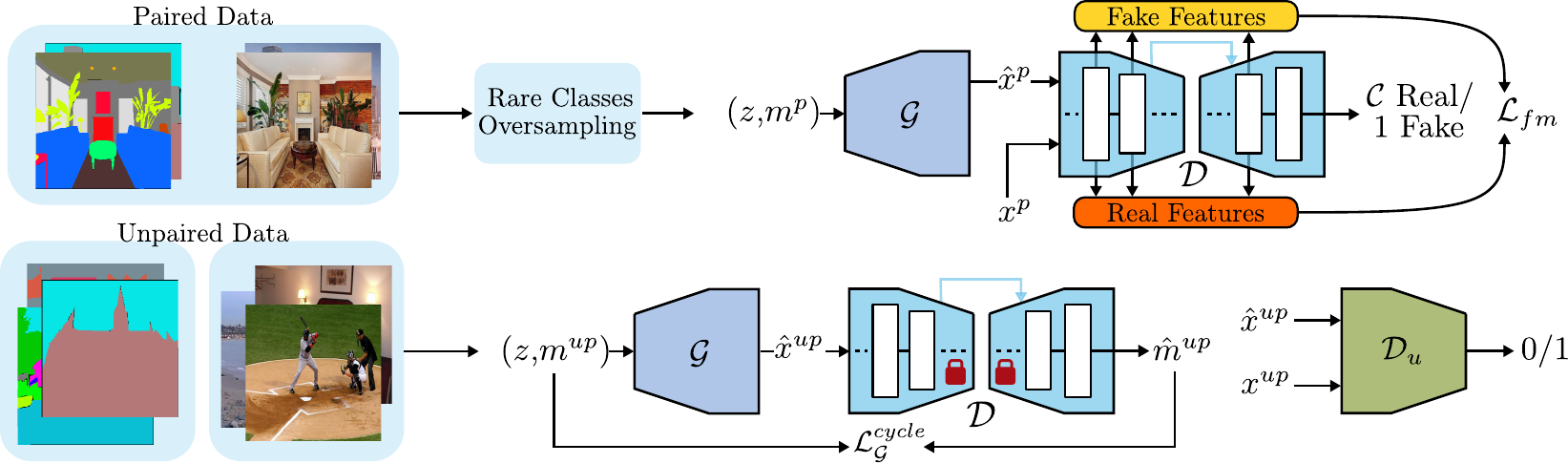}
    \end{center}
    \vspace{-0.5em}
    \caption{\small The generator is trained to reconstruct the labels of the unpaired data as an additional pretext task. The reconstruction network is chosen to be the U-Net conditional discriminator $\mathcal{D}$ (with frozen weights) on the paired data and operates as a reverse mapping generator on the unpaired data. Paired data containing rare semantic classes are over-sampled in the first training phase. }
    \label{fig:framework}
    \vspace{-1.5em}
\end{figure*}

We propose a semi-paired framework featuring a generator $\Gen$, and 2 discriminators: an unconditional discriminator $\Du$ and a U-Net~\cite{Ronneberger2015UNetCN} shaped conditional discriminator, $\Dis$. $\Gen$ is our designed waveletSPADE generator in \cite{usisicassp} while $\Du$ is the SWAGAN discriminator~\cite{Gal2021SWAGANAS, usisicassp}. $\Dis$ attempts to model the joint probability distribution of images and labels $\mathcal{P}(x,m)$ from the small amount of paired data at its disposal, while $\Du$ models the marginal probability distributions of images alone, $\mathcal{P}(x)$. On the unpaired data, $\Dis$ tries to push the generator to produce images that are constrained by the learned joint probability distribution by operating as the reverse mapping network ($\Gen_{rev}$) in a cycle loss (the generated images from unpaired data $\hat{x}^{up}$ are segmented back to the input masks $m^{up}$). Thus, reverse mapping and conditional discrimination are learned in a synergistic way and shared by the same network (Figure ~\ref{fig:framework}). Similar to USIS~\cite{usisicassp}, $\Du$ is essential to prevent trivial identity mappings which can be realised by only minimizing the cycle loss on unpaired data. The motivation of this approach is to introduce more cooperation between the paired and unpaired tasks by transferring the knowledge learned on paired data through conditional discrimination to unpaired data, so that the generator can learn a reverse mapping based on $\mathcal{P}(x,m)$ and not only on $\mathcal{P}(x)$ as in \cite{zhu2017unpaired, lee2018diverse, yi2017dualgan}. However, since the unpaired data is more abundant than the paired data, $\Dis$ might give a bigger weight to minimizing the cycle loss and deviate from representing the true distribution of real data by reaching an undesired local minimum. For this reason, we freeze the parameters of $\Dis$ when the segmentation cycle loss on unpaired data is backpropagated (indicated by red lock in Figure~\ref{fig:framework}).


\vspace{-1.5em}
\subsection{Training algorithm}
\vspace{-0.75em}
During each training iteration, we sample a batch from $\DataPaired$ and a batch from $\DataUnpaired$. The training first starts on the batch from $\DataPaired$ as $\Dis$ tries to segment real paired images $x^p$ to their semantic map $m^p$, and to segment fake images $\hat{x}^p$ to a fake class $c=\Cl + 1$. The loss function of $\Dis$ is expressed as: 
\begin{equation}
    \begin{aligned}
    \nonumber
    \mathcal{L}_{\Dis} = &- \mathbb{E}_{(x^p, m^p)} \left[ \sum_{c=1}^{C} \alpha_c \sum_{h,w}^{H \times W} m^p_{h,w,c} \log(\Dis(x^p)_{h,w,c}) \right] \\
    &- \mathbb{E}_{(z, m^{p})} \left[ \sum_{h,w}^{H \times W} \log(\Dis(\hat{x}^p)_{h,w,c=\Cl +1})) \right] \\  
    \alpha_{c} =& \frac{H \times W}{ \sum_{h,w}^{H \times W} \mathbb{E}_{m} \left[ \mathbbm{1}[m_{h,w,c}] \right] }\end{aligned}
\end{equation}

On the other hand, $\Gen$ learns an adversarial objective by trying to segment the fake images into their original semantic classes. We also add a feature matching loss $\mathcal{L}_{fm}$ on each layer of $\Dis$ (except the final layer) and a VGG-loss~\cite{wang2018high}. Although in \cite{schonfeld2021you} the feature matching loss is unnecessary, we find that it is more essential in the semi-paired setting, especially for small $r$. When the number of paired examples is low, the feature matching loss can enforce a more consistent feature representation between real and fake images inside $\Dis$. The loss function for $\Gen$ on the paired data becomes:
\begin{equation}
    \begin{aligned}
    \nonumber
    \mathcal{L}_{\Gen}^{sup} = &- \mathbb{E}_{(z, m^p)} \left[ \sum_{c=1}^{C} \alpha_c \sum_{h,w}^{H \times W} m^p_{h,w,c} \log(\Dis(\hat{x}^p)_{h,w,c}) \right] \\
    & + \mathbb{E}_{(x^p, m^p)} \left[ \sum_{l=1}^{L-1} (\Dis^l(x^p) - \Dis^l(\hat{x}^p))^2 \right] +\mathcal{L}_{VGG}, \\ 
    \end{aligned}
\end{equation}
where $\Dis^l$ is the $l$-th layer of $\Dis$ and $\mathcal{L}_{VGG}$ is the VGG content loss in \cite{wang2018high}. Afterwards, the training continues on the unpaired batch, where $\Gen$ tries to fool $\Dis$ into segmenting $\hat{x}^{up}$ back to $m^{up}$ via a cycle loss $\mathcal{L}_{\Gen}^{cycle}$. The output fake class ($\Cl + 1$) from $\Dis$ is ignored. During the backpropagation of this loss, the parameters of $\Dis$ are not updated. At the same time, $\Gen$ and $\Du$ are trained adversarially with the traditional GAN loss~\cite{Karras2020AnalyzingAI}. The unsupervised losses can be expressed as follows: 
\begin{equation}
    \begin{aligned}
    \nonumber
    \mathcal{L}_{\Gen}^{uns} = &- \mathbb{E}_{(z, m^{up})} \left[ \sum_{c=1}^{C} \alpha_c \sum_{h,w}^{H \times W} m^{up}_{h,w,c} \log(\Dis_{\mathbf{fr}}(\hat{x}^p)_{h,w,c}) \right] \\
    & - \mathbb{E}_{(z, m^{up})} \left[ log(\Du(\hat{x}^{up}))) \right], \\     \end{aligned}
\end{equation}
where $\Dis_{\mathbf{fr}}$ denote $\Dis$ with frozen parameters. 
\begin{equation}
    \begin{aligned}
    \nonumber
    \mathcal{L}_{\Du} = &- \mathbb{E}_{(z, m^{up})} \left[ 1- log(\Du(\hat{x}^{up})) \right] - \mathbb{E}_{x^{up}} \left[ log(\Du(x^{up})) \right]. \\ \end{aligned}
\end{equation}
While it is possible to apply $\mathcal{D}_u$ on $\DataPaired$ as well, we notice no increase in performance when we do.

\vspace{-1em}
\subsection{Sampling of Rare Classes}
A usual problem in SIS is class imbalance: some classes are only present in a few samples in the dataset; hence, the generator does not observe sufficient instances of these classes. The problem is aggravated in the unpaired and semi-paired settings, especially when the unpaired samples outnumber the paired samples. To prevent $\Gen$ from ignoring the rare classes, images featuring these rare classes in $\DataPaired$ are given a higher probability of being sampled in the first half of the training. Specifically, we propose to assign each image $x^p_i$ a sampling probability $p_i$ such that images with rare classes are given a higher sampling probability. For each image, we calculate its share of classes in the dataset. The share of a class $c$ in $x^p_i$ is calculated as the ratio of the number of pixels with class $c$ in the image ($n^c_i$) to the total number of pixels with class $c$ in the paired dataset ($n^c_{\DataPaired}$). An image containing a rare class will have a high share of this class, and will thus be oversampled. Note the sampling probability $p_i$ is also biased towards smaller classes as well, as they can be considered rare in terms of the number of pixels covered per image by the class. We obtain $p_i$ by first calculating $\hat{p}_i$ as:
\begin{equation}
    \begin{aligned}
    \nonumber
    \small
    \hat{p}_i = \sum_{c=1}^{\Cl} \mathbbm{1}^c_i \frac{n^c_i}{n^c_{\DataPaired}},\\ \end{aligned}
\end{equation}
where $\mathbbm{1}^c_i$ is a boolean indicating whether $m^p_i$ contains at least one pixel with class $c$. All $\hat{p}_i$ are then divided by their sum to obtain probabilities $p_i$ in $[0,1]$. 

\begin{figure*}[t]
    \captionsetup[subfloat]{position=top, labelformat=empty, skip=0pt}
    \centering
    \begin{minipage}{0.85\textwidth}
    \subfloat[Label]{\includegraphics[width=  0.198\textwidth, height=  0.12\textwidth]{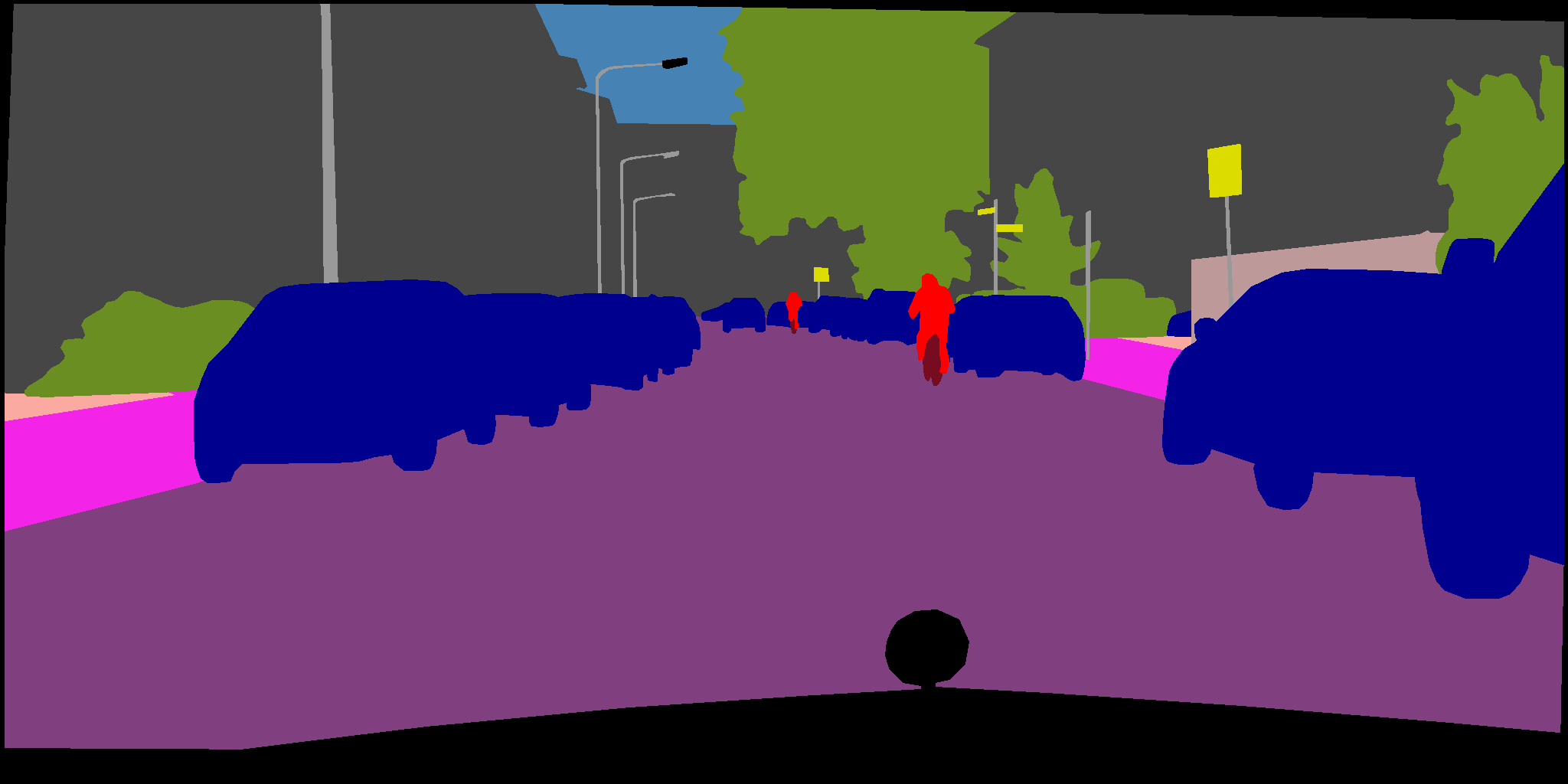}} \hfill
    \subfloat[OASIS \cite{schonfeld2021you}]{\includegraphics[width=  0.198\textwidth, height=  0.12\textwidth]{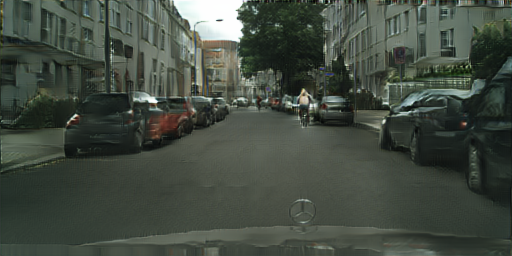}} \hfill
    \subfloat[TCR \cite{tcr}]{\includegraphics[width=  0.198\textwidth, height=  0.12\textwidth]{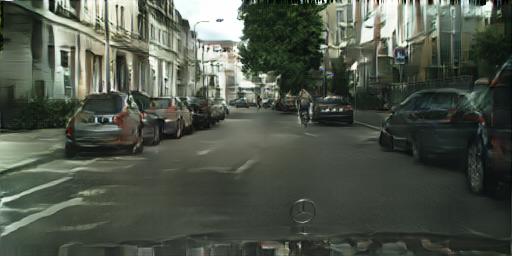}} \hfill
    \subfloat[Ours]{\includegraphics[width=  0.198\textwidth, height=  0.12\textwidth]{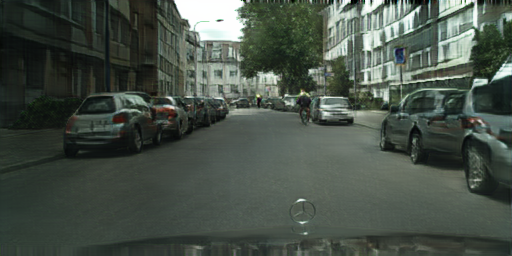}} \hfill
    \subfloat[Groundtruth]{\includegraphics[width=  0.198\textwidth, height=  0.12\textwidth]{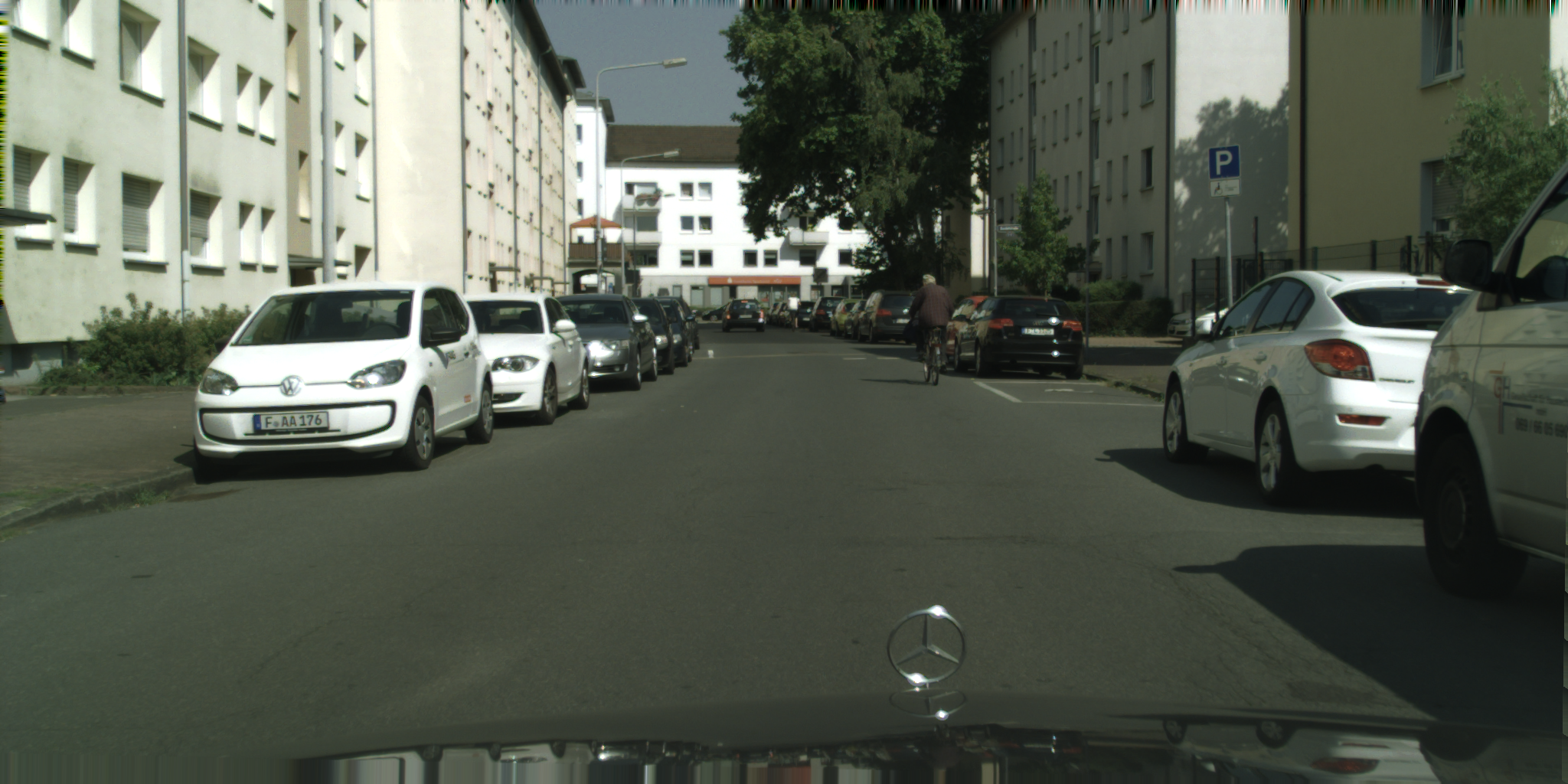}}
    
    \vspace{-1em}
    \subfloat[]{\includegraphics[width=  0.198\textwidth, height=  0.12\textwidth]{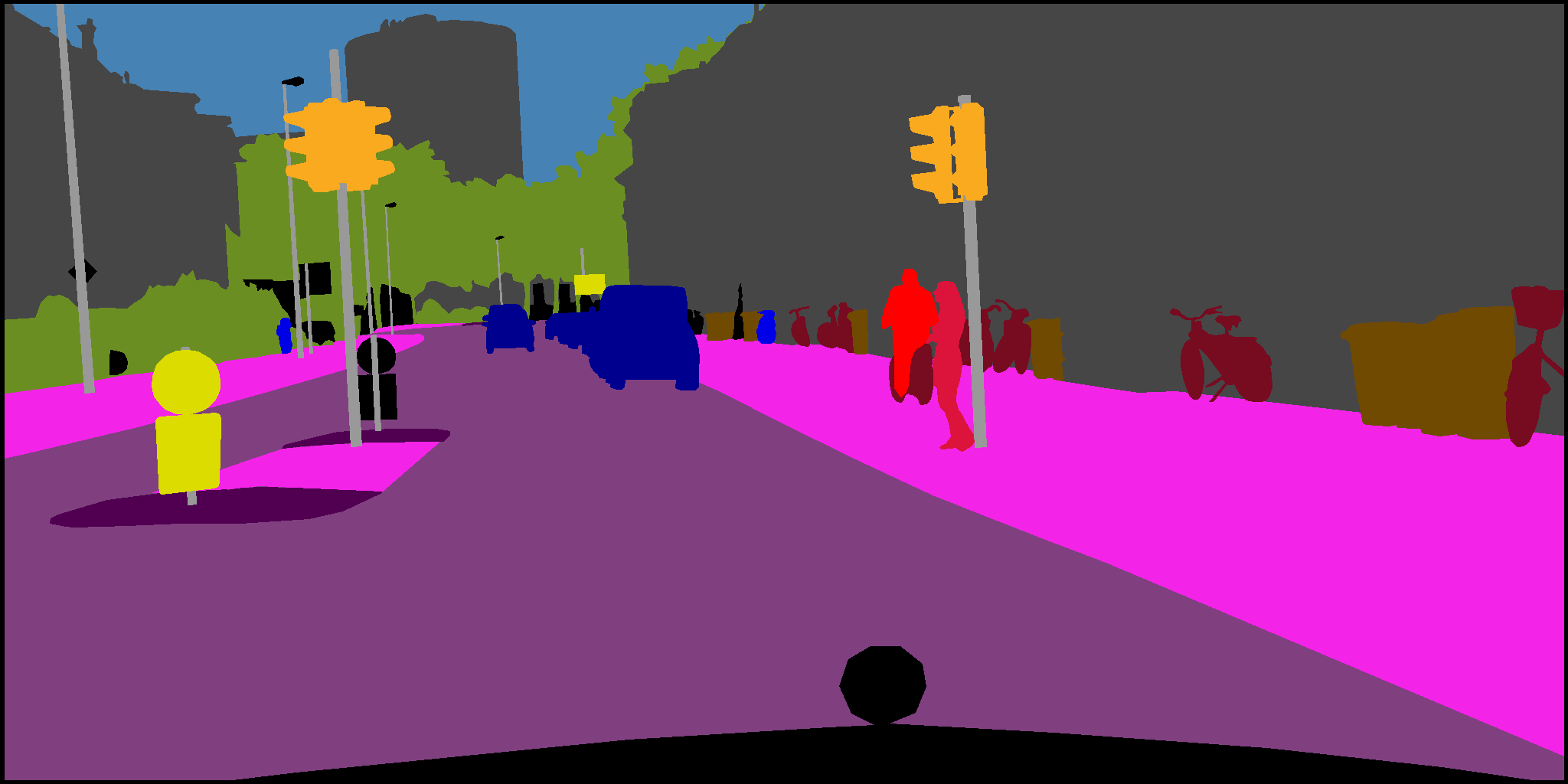}} \hfill
    \subfloat[]{\includegraphics[width=  0.198\textwidth, height=  0.12\textwidth]{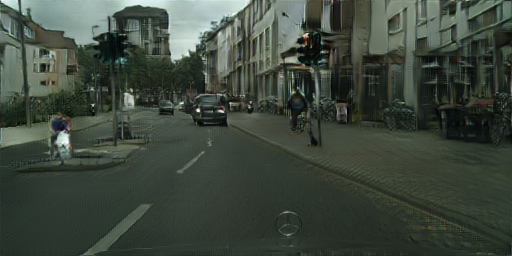}} \hfill
    \subfloat[]{\includegraphics[width=  0.198\textwidth, height=  0.12\textwidth]{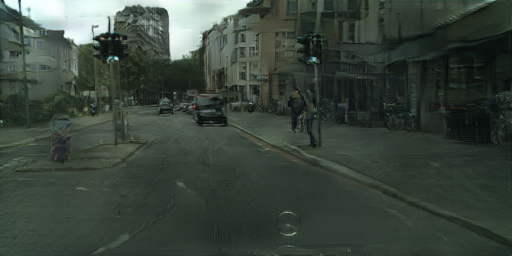}} \hfill
    \subfloat[]{\includegraphics[width=  0.198\textwidth, height=  0.12\textwidth]{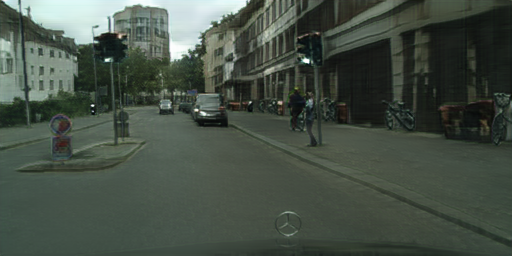}} \hfill
    \subfloat[]{\includegraphics[width=  0.198\textwidth, height=  0.12\textwidth]{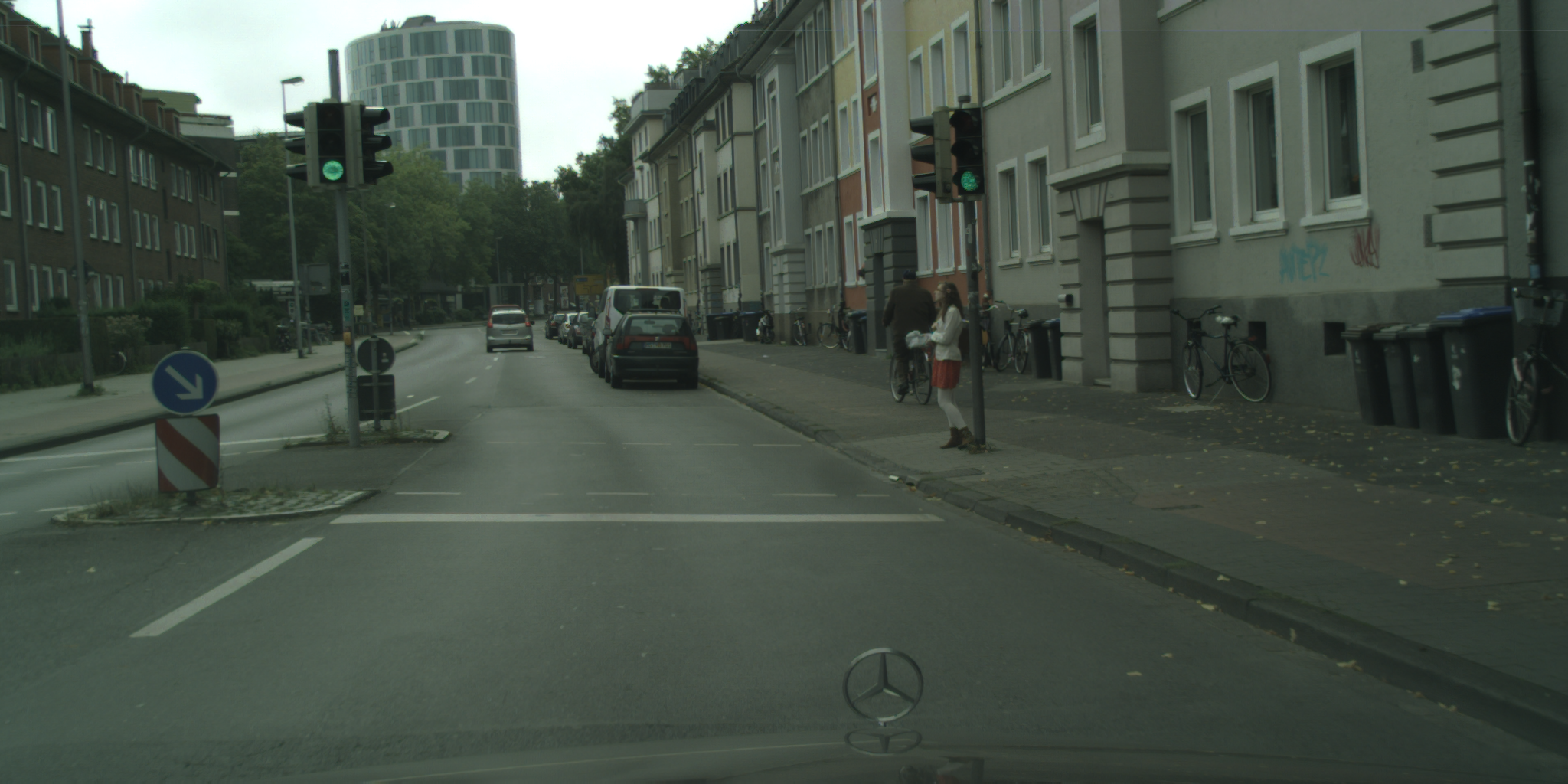}}
    \vfill
    \end{minipage}
   \caption{Comparison on Cityscapes ($r=20\%$). Our model generates more realistic results, especially for smaller classes.}
   \vspace{-1.5em}
   \label{fig:results}
\end{figure*}

\vspace{-1em}
\section{Experiments}
\vspace{-1em}
\label{sec:experiments}
\textbf{Datasets and Metrics.} Similar to OASIS~\cite{schonfeld2021you} and USIS~\cite{usisicassp}, we conduct experiments on 3 standard benchmarks:  Cityscapes \cite{cordts2016cityscapes}, ADE20K~\cite{zhou2017scene} and COCO-stuff~\cite{caesar2018coco}. We use the same image resolution in \cite{usisicassp}. Since in the practical case, $N_{up} > N_p$, we test the proposed framework on multiple $r$, where $r<50\%$. On Cityscapes and ADE20K, we test with $r = 10, 20, 40, 60, 100 \%$. On COCO-stuff, we try more challenging scenarios with very small $r$; $r=0.5, 1.0, 5.0\%$. COCO-stuff and ADE20K have significantly more samples than Cityscapes ($120k$ and $20k$ versus $3k$). We use a batchsize of 8 and a learning rate is 0.0001 in all experiments. Similar to \cite{schonfeld2021you, park2019semantic, usisicassp}, we measure the FID~\cite{heuselttur2017} to assess the diversity and quality, and mean Intersection-over-Union (mIoU) of the generated images compared to the groundtruth labels to assess semantic alignment and quality.

\noindent \textbf{Baselines.} To our knowledge, TCR~\cite{tcr} is the only semi-supervised paradigm for I2I. It was only applied to image denoising, colorization and super-resolution and has shown that by combining $\DataPaired$ and $\DataUnpaired$, it outperforms the supervised model trained on the same $\DataPaired$ only. Similarly, we compare our model with: the supervised OASIS trained on $\DataPaired$, the semi-supervised TCR ($\DataPaired \cup \DataUnpaired$), the fully unpaired USIS, the fully paired models: SPADE~\cite{park2019semantic}, CC-FPSE~\cite{liu2019learning} and OASIS. For a fair comparison, as TCR was not applied to SIS, we use the generator and the supervised adversarial loss from OASIS for TCR, so that TCR has a similar learning capacity. For each value of $r < 100\%$ in Tables~\ref{tab:results} and \ref{tab:coco}, we first split the dataset into 2 fixed splits: $\DataPaired$ and $\DataUnpaired$. OASIS is trained on $\DataPaired$, TCR and our model are trained on $\DataPaired \cup \DataUnpaired$.

\vspace{-1em}

\section{Results}
\label{sec:results}
\vspace{-1em}
We show quantitative results in Table~\ref{tab:results} and \ref{tab:coco}, and qualitative results in Figure~\ref{fig:results}. The proposed model outperforms USIS, OASIS and TCR on all 3 datasets. Our framework is significantly better in the more challenging settings when $N_p$ is small (empirically $< 2000$), which corresponds to $r = 10,20,40\%$ on Cityscapes  and to $r=0.5, 1\%$ on COCO-stuff. When $N_p$ is high ($> 2000$), we still outperform both in FID while resulting in similar or slightly higher mIoU. Moreover, our model sometimes generates better results than OASIS and TCR while using less paired data than both of them (for example, Ours with $r=20\%$ outperforms OASIS and TCR with $r=40\%$ on Cityscapes). It can even reach better results than some fully supervised methods while using less paired data (Ours with $r=20\%$ versus SPADE, Ours with r=$40\%$ versus CC-FPSE on Cityscapes, Ours with $r=40\%$ versus SPADE on ADE20K). Finally, we demonstrate the strength of the proposed approach in the fully paired setting as it outperforms the supervised baselines. Note that sometimes our model can slightly produce a worse FID with higher values of $r$ (Ours with $r=20\%$ versus Ours with $r=40\%$). This is because the model might easily imitate the patch distribution of real images (low FID) when $r$ is small by ignoring some classes in the semantic map resulting in undesirable semantic misalignments. However, when $r$ is higher, the model learns these classes (high mIoU) but might not always generate a realistic texture for these challenging classes, hence the slightly worse FID.

\begin{table}[t!]

	 %

	\setlength{\tabcolsep}{0.2em}
	\renewcommand{\arraystretch}{0.95}
	\centering
	
    \scalebox{0.85}{\begin{tabular}{l|c|cc|cc}
    \multirow{2}{*}{ Method } & \multirow{2}{*}{$r$} & \multicolumn{2}{c|}{ Cityscapes} &  \multicolumn{2}{c}{ ADE20K} 
    \tabularnewline
      &   & FID$\downarrow$  &  mIoU$\uparrow$ &   FID$\downarrow$  &  mIoU$\uparrow$   \tabularnewline
    	
    \hline 
    	
    \small{} USIS &  0.0	& \textcolor{darkblue}{50.1}  &  \textcolor{darkblue}{42.3} &  \textcolor{darkblue}{34.5} &  \textcolor{darkblue}{16.9}  \tabularnewline 
    	
     \hline
     
     \small{} OASIS	& \multirow{3}{*}{$10$} & 66.1  & 57.8  & 45.9  & 34.4     \tabularnewline
    	
    \small{} TCR	&  &  98.4 &  56.4 & 47.8  & 32.3   \tabularnewline
    	
    \small{} Ours	&  & \textbf{48.9} &  \textbf{60.4} &  \textbf{37.6} &  \textbf{36.9}      \tabularnewline
    	
    	
    	
    	
    \hline
    
    \small{} OASIS	& \multirow{3}{*}{$20$} &  57.3 &  60.8 &  38.7 &  39.1   \tabularnewline
    	
    \small{} TCR	&  &  70.3 &  61.1 &  37.6 &  37.6   \tabularnewline
    	
    \small{} Ours	&  & \textbf{45.6} &  \textbf{64.8} &  \textbf{34.9} &  \textbf{40.8}     \tabularnewline
    	
    	
    	
    	
    \hline
    
    \small{} OASIS	& \multirow{3}{*}{$40$} & 49.9 & 62.58 &  37.2 &  38.3  \tabularnewline
    	
    \small{} TCR	&  &  59.7 &  64.5 &  34.5 &  40.6    \tabularnewline
    	
    \small{} Ours	&  & \textbf{46.8} &  \textbf{66.6} &  \textbf{31.6} &  \textbf{41.3}    \tabularnewline
    
    \hline 
    
    \small{} OASIS	& \multirow{3}{*}{$60$} & 49.9 &  62.9 &  36.8 &  42.1  
    \tabularnewline
    	
    \small{} TCR	&  &  49.6 &  66.7 & 34.9 &  42.0 
    
    \tabularnewline
    	
    \small{} Ours	&  &  \textbf{47.5} &  \textbf{67.4} & \textbf{33.5} &  \textbf{43.2} 
    \tabularnewline
    	
    \hline
    
    \small{} SPADE	& \multirow{5}{*}{100}  & 71.8  & 62.3  & 33.9  & 38.5   \tabularnewline 
    
    \small{} CC-FPSE &  & 54.3  & 65.5  & 31.7  & 43.7    \tabularnewline
    
    \small{} OASIS	&   &  47.7 &  69.3 &  28.3 &  48.8 \tabularnewline 
    
    \small{} TCR	&   & 52.4  & 69.8  & 30.7  & 48.2  \tabularnewline 
    
    \small{} Ours	&   &  \textbf{46.4} & \textbf{71.1}  & \textbf{27.4}  & \textbf{49.2}  \tabularnewline 
    
    
	
	\end{tabular}}
	\caption{Results on Cityscapes and ADE20K datasets}
	\label{tab:results}
	\vspace{-2.5em}
\end{table}

To show the importance of the proposed contributions, we ablate different parts of the framework and report the performance on the Cityscapes datasets for $r=20\%$ ($N_p = 600$) in Table~\ref{tab:ablation}. In each model, only one part of the proposed framework is removed while the rest is kept unchanged. We compare against OASIS trained on the same $N_p$ paired samples only (first row). One might expect that using more data ($\DataUnpaired$) must naturally increase the performance but we show that this is \textit{not} the case in rows A and B as SIS is a challenging task, that requires a careful design. The model in row A has 2 different networks, $\Dis$ and $G_{rev}$, which are trained separately and can be considered as a straightforward aggregation of OASIS (on $\DataPaired$) and USIS (on $\DataUnpaired$). We observe that the mIoU significantly lags behind the supervised baseline, showing that more data and/or bigger model capacity do not readily lead to better performance. In B, we only unfreeze $\Dis$ during $\mathcal{L}_{unsup}$. While both FID and mIoU improve relative to the previous model, mIoU remains lower than OASIS, highlighting the importance of freezing $\Dis$. Finally, we study the effect of removing $\mathcal{L}_{fm}$, $\mathcal{L}_{VGG}$, and the proposed rare classes sampling. We observe that without the proposed rare classes sampling, the mIoU drops, while the absence of either $\mathcal{L}_{VGG}$ or $\mathcal{L}_{fm}$ affect both FID and mIoU.


\begin{table}[t!]
\centering
    \adjustbox{width=1.0\linewidth}{\begin{tabular}{l|c|cc}
         Method & Data & FID & mIoU  \\ 
         \hline
        OASIS & $\DataPaired$                                & 57.3 & 60.8 \\
        \hline
        A- OASIS + USIS      & $\DataPaired \cup \DataUnpaired$                   & 53.3 & 49.9 \\
        B- Shared unfrozen $\Dis$ & $\DataPaired \cup \DataUnpaired$               & 47   & 56.8 \\
        C- w/o $\mathcal{L}_{fm}$  &    $\DataPaired \cup \DataUnpaired$          & 49.8 & 62.0 \\
        D- w/o $\mathcal{L}_{VGG}$  &  $\DataPaired \cup \DataUnpaired$           & 51.0 & 61.9 \\
        E- w/o rare classes sampling & $\DataPaired \cup \DataUnpaired$           & 48.1 & 61.8 \\
        F- w/ rare classes sampling in 2nd training phase & $\DataPaired \cup \DataUnpaired$ & 46.5 & 64.7  \\ 
        \hline
        Ours (complete)               & $\DataPaired \cup \DataUnpaired$       & \textbf{45.6} & \textbf{64.8} \\
        \end{tabular}}
\caption{Ablation study on Cityscapes with $r=20\%$.}
\label{tab:ablation}
\vspace{-1em}
\end{table}
\begin{table}[t!]
    \centering
    \adjustbox{width=0.95\linewidth}{\begin{tabular}{c|cc|cc|cc}
        \multirow{2}{*}{Method} & \multicolumn{2}{c|}{$r=0.5\%$} & \multicolumn{2}{c|}{$r=1.0\%$} & \multicolumn{2}{c}{$r=5\%$} \\
        & FID$\downarrow$  &  mIoU$\uparrow$ & FID$\downarrow$  &  mIoU$\uparrow$ & FID$\downarrow$  &  mIoU$\uparrow$ \\
        \midrule
        USIS ($r=0\%$) & 28.6 & 13.4 & 28.6 & 13.4 & 28.6 & 13.4 \\
        \hline 
        OASIS & 180.5 & 0.57 & 64.5 & 25.7 & 34.1 & \textbf{39.1} \\
        TCR   & 111.1 & 11.7 & 74.4 & 21.4 & 34.2 & 35.6 \\
        Ours  & \textbf{50.3} & \textbf{28.1} & \textbf{41.5} & \textbf{32.1} & \textbf{28.9} & 38.3 \\
        \midrule
        SPADE ($r=100\%$) & 22.6 & 37.4 & 22.6 & 37.4 & 22.6 & 37.4 \\
        OASIS ($r=100\%$) & 17.0 & 44.1 & 17.0 & 44.1 & 17.0 & 44.1 \\
    \end{tabular}}
    \caption{Results on COCO-stuff in low-data regime}
    \label{tab:coco}
    \vspace{-2em}
\end{table}

\vspace{-1em}
\section{Conclusion}
\vspace{-0.7em}
\label{sec:conclusion}
We propose the first semi-paired framework for SIS, which outperforms state-of-the-art unsupervised and semi-supervised GANs on 3 challenging datasets and even some fully supervised models, all while using much smaller supervision ratios. Moreover, we show that the proposed model improves upon supervised frameworks when using all available labeled samples. This was achieved through 2 novel contributions: a framework with 2 discriminators, where the supervised discriminator operates as a reverse mapping generator on unpaired data, and a rare classes sampling algorithm. Finally, the proposed method does not preclude the use of TCR or additional pretext tasks to further improve the quality.   
\vspace{-1em}
\section*{\centering \normalsize Acknowledgement}
\label{sec:ack}
\vspace{-1em}
\setstretch{0.6}
{\small The research leading to these results is funded by the German Federal Ministry for Economic Affairs and Energy within the project "AI Delta Learning".  The authors would like to thank the consortium for the successful cooperation.}

\vspace{-1em}


\clearpage
\begingroup
\setstretch{0.82}
\printbibliography
\endgroup

\end{document}